%% file: iclr2025_conference.tex
\documentclass{article} 
\usepackage{iclr2025_conference,times}

\input{math_commands.tex}

\usepackage{hyperref}
\usepackage{url}
\usepackage{graphicx}
\usepackage{enumitem}

\usepackage{amsmath}

\usepackage{booktabs}
\usepackage{xcolor}
\usepackage{soul} 
\usepackage{colortbl} 
\usepackage{caption}  
\usepackage{tabularx}

\usepackage{graphicx}
\usepackage{subcaption}

\definecolor{mygreen}{RGB}{34,139,34}
\definecolor{myred}{RGB}{220,20,60}
\definecolor{lightblue}{RGB}{230,245,250}

\newcommand{\up}[1]{\textcolor{mygreen}{\tiny +{#1}}}

\newcommand{\abs}[1]{\lvert #1 \rvert}

\title{G-MemLLM: Gated Latent Memory Augmentation for Long-Context Reasoning in Large Language Models}

\author{Xun Xu\\
Department of Computer Science\\
Fudan University\\
\texttt{23307130122@m.fudan.edu.cn} \\
}

\iclrfinalcopy 
\begin{document}

\maketitle
 
\begin{abstract}
Large Language Models (LLMs) have demonstrated remarkable capabilities in natural language understanding, yet they remain constrained by the finite capacity of their context windows and the inherent difficulty of maintaining long-term factual consistency during multi-hop reasoning. While existing methods utilize context compression or recurrent tokens, they often suffer from ``context rot'' or the dilution of information over long horizons. In this paper, we propose \textbf{G-MemLLM}, a memory-augmented architecture that integrates a frozen LLM backbone with a trainable \textbf{Latent Memory Bank}. Our key innovation is a GRU-style gated update logic that allows the model to selectively update, preserve, or overwrite latent memory slots, preventing the vanishing gradients of knowledge common in recurrent systems. We evaluate G-MemLLM across scales, from GPT-2 (124M) to Llama 3.1 (8B), on the HotpotQA and Zero-Shot Relation Extraction (ZsRE) benchmarks. Our results demonstrate that G-MemLLM significantly enhances multi-hop reasoning and relational precision, achieving a 13.3\% accuracy boost on ZsRE for Llama 3.1-8B, and it also yields improvements across model scales, boosting Answer F1 by 8.56 points for GPT-2 and increasing Supporting Fact F1 by 6.89 points for Llama 3.1-8B on HotpotQA.
\end{abstract}

\section{Introduction}

The evolution of Large Language Models (LLMs) has been defined by a constant tension between model capacity and context management. While modern models possess vast parametric knowledge, their ability to synthesize information across disparate documents or maintain factual integrity over extended interactions is often limited by the quadratic complexity of the transformer's attention mechanism. This limitation is particularly evident in tasks requiring multi-hop reasoning, such as HotpotQA \citep{yang2018hotpotqa}, where a model must bridge multiple facts to reach a conclusion, or in relational knowledge extraction \citep{levy2017zero}, where precise entity mapping is required.

Recent attempts to extend the functional context of LLMs have generally followed two paths: context compression and recurrent state-passing. Context compression methods, such as Gist Tokens \citep{mu2023learning}, attempt to condense prefixes into a smaller set of virtual tokens. However, these often lead to information bottlenecks where fine-grained details are lost. On the other hand, recurrent architectures like the Recurrent Memory Transformer (RMT) \citep{bulatov2022recurrentmemorytransformer} pass hidden states across segments but struggle with vanishing knowledge as information is progressively overwritten by incoming noise in the sequence.

A central challenge in these systems is \textit{gating}: the ability of a model to decide which information is transient and which is of long-term utility. Without an explicit mechanism to manage the lifecycle of a memory slot, latent representations eventually converge toward a mean, losing the specificity required for complex reasoning.

In this work, we introduce \textbf{G-MemLLM}, a gated latent memory augmented framework designed to provide LLMs with a persistent and selectively updatable working memory. G-MemLLM decouples linguistic processing from knowledge retention by utilizing a frozen LLM backbone paired with a trainable Latent Memory Bank. Unlike previous passive memory pools, G-MemLLM employs a differentiable, GRU-style gated update mechanism. This gate allows the model to dynamically regulate the flow of information into its latent slots, ensuring that bridge entities in multi-hop queries are preserved until they are no longer needed.

\begin{figure}[htbp]
     \centering
     \begin{subfigure}[b]{0.48\textwidth}
         \centering
         \includegraphics[width=\textwidth]{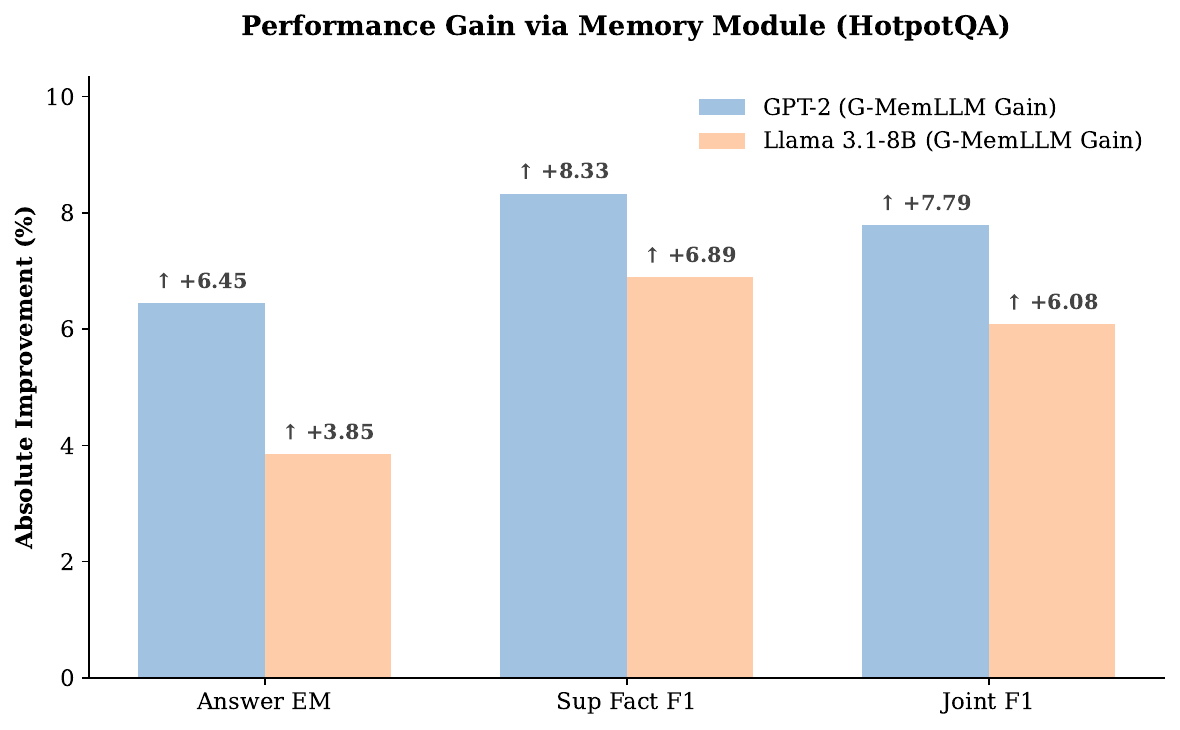}
         \caption{Improvement on HotpotQA}
         \label{fig:hotpot_delta}
     \end{subfigure}
     \hfill 
     \begin{subfigure}[b]{0.48\textwidth}
         \centering
         \includegraphics[width=\textwidth]{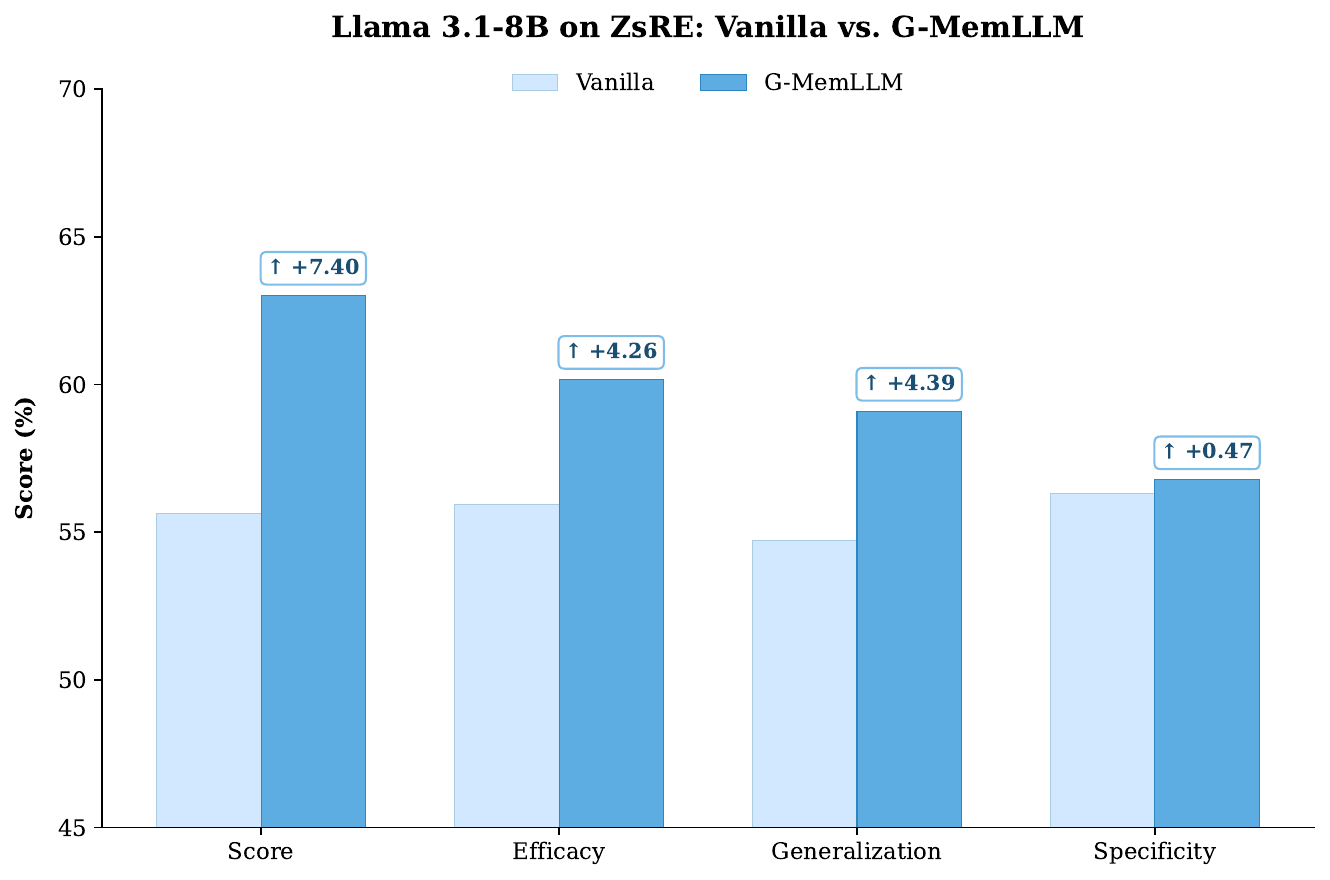}
         \caption{Comparison on ZsRE}
         \label{fig:zsre_blue}
     \end{subfigure}
     
     \caption{\small Performance enhancement brought by the G-MemLLM memory module. (a) illustrates the absolute improvement across models, while (b) shows the direct comparison between Vanilla and G-MemLLM on specific metrics.}
     \label{fig:overall_results}
\end{figure}

We demonstrate the efficacy of G-MemLLM through a rigorous scaling analysis. By testing our module on both GPT-2 (124M) and Llama 3.1 (8B), we show that our architecture is not only effective for small-scale models but still remains efficient as the base model's capacity increases. Our contributions are as follows:

\begin{itemize}
    \item We propose the G-MemLLM architecture, featuring a Latent Memory Bank with a gated update logic that prevents memory drift and information overwrite.
    \item We introduce a composite training objective that combines sparsity and entropy losses to ensure efficient and diverse memory slot utilization.
    \item We provide an empirical evaluation on \textbf{HotpotQA} \citep{yang2018hotpotqa} and \textbf{ZsRE} \citep{levy2017zeroshotrelationextractionreading}, showing that G-MemLLM bridges the performance gap for small models and significantly extends the reasoning capabilities of large-scale pre-trained models.
\end{itemize}

\section{Related Work}

\subsection{Context Compression and Information Bottlenecks}

The primary bottleneck for processing long sequences in Transformer-based architectures is the quadratic growth of the KV-cache. To address this, several works have explored information distillation through compression. \cite{mu2023learning} introduced Gist Tokens, which utilize a soft-token bottleneck to condense long prompt prefixes into a few learnable virtual tokens. Building on this, Recurrent Context Compression (RCC) \citep{huang2024recurrentcontextcompressionefficiently} demonstrated that frozen encoders can recursively compress million-token contexts into compact representations. While these methods successfully reduce inference latency, they often suffer from context rot \citep{hong2025context}—a phenomenon where the density of information becomes too high for the decoder to reconstruct specific facts, leading to non-uniform performance degradation in long-horizon tasks.

\subsection{Memory Tokens and Recurrent State-Passing}

Another trajectory focuses on maintaining a "working memory" through persistent tokens that flow across segments. \cite{bulatov2022recurrent} pioneered this with the Recurrent Memory Transformer (RMT), which passes memory tokens between segments to maintain global context. More recently, the M+ framework \citep{wang2025mplus} extended this concept by integrating a co-trained retriever that expands the usable latent-space memory to over 160k tokens. However, as noted in the LOCCO benchmark \citep{jia-etal-2025-evaluating}, simple recurrent passing often leads to vanishing gradients of knowledge, where information from the beginning of a sequence is progressively overwritten or diluted by noise. These systems lack a robust mechanism to differentiate between transient context and permanent knowledge.

\subsection{Dynamic Latent Memory and Gated Architectures}

The most recent shift involves moving memory entirely into the latent space to simulate human-like cognitive architectures. \cite{wang2024memoryllm} introduced MemoryLLM, which utilizes a large-scale memory pool for self-updatable knowledge. Our G-MemLLM aligns with this philosophy but introduces a critical architectural innovation: the GRU-style gated update logic. While previous models like ERMAR \citep{alselwi2025longcontextmodelingranked} rely on passive accumulation or ranking, G-MemLLM uses a differentiable gate ($g$) to manage information overwrite. This allows for the precise integration of multi-hop facts while maintaining the stability of the frozen LLM backbone.

\begin{figure}[htbp]
  \centering
  \includegraphics[width=1\linewidth]{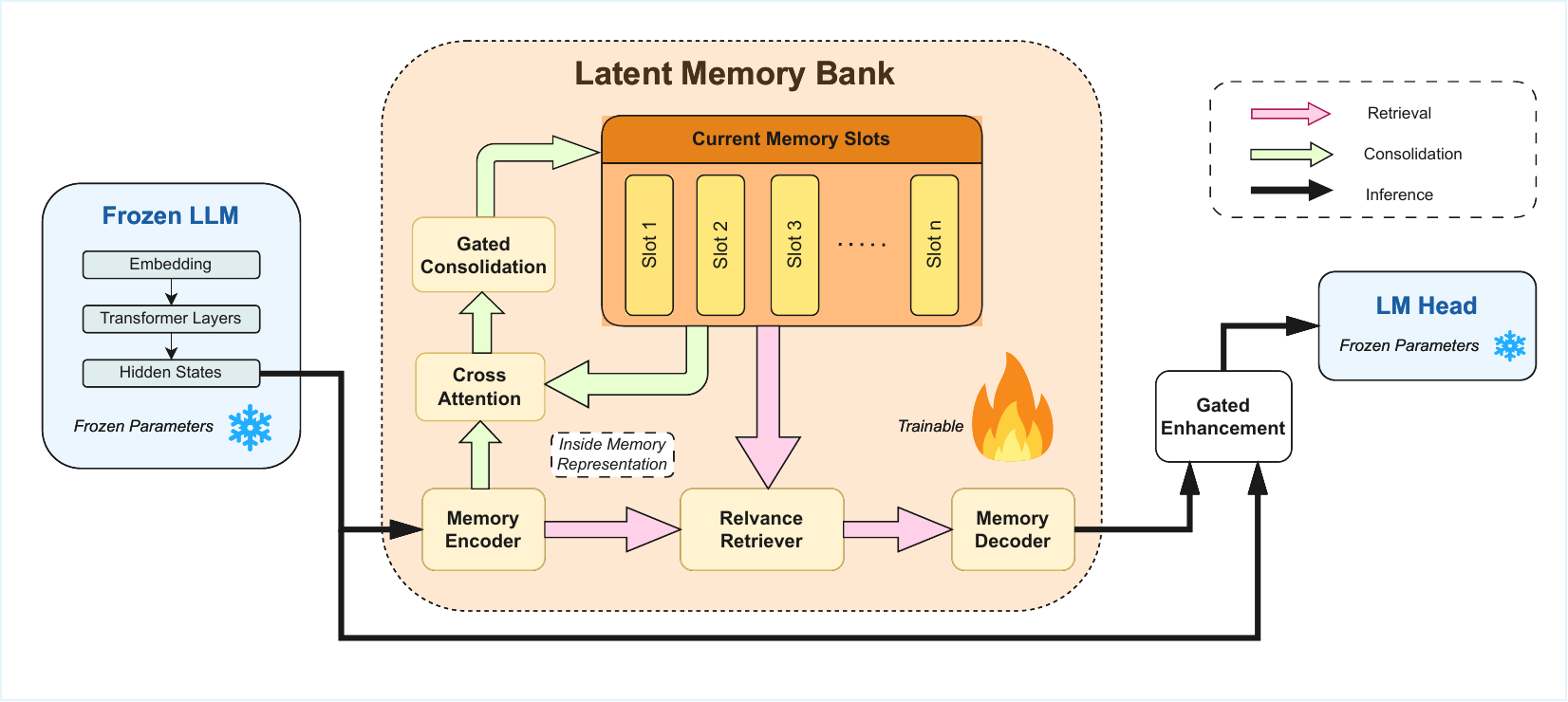}
  \caption{Overview of G-MemLLM architecture.}
  \label{fig:example}
\end{figure}

\section{Method}

The G-MemLLM architecture is designed to bridge the gap between static, pre-trained knowledge and dynamic, task-specific context. It consists of two primary systems: a frozen LLM backbone that provides linguistic features and a trainable latent memory bank that manages persistent state.

\subsection{Latent Memory Bank}

The latent memory bank is the central working memory of the agent. It manages a fixed number of learnable memory slots, $M \in \mathbb{R}^{S \times D_m}$ (where $S$ is the number of slots and $D_m$ is the memory dimension).

\textbf{Memory encoder and decoder} \quad
These sub-networks act as translators. The encoder compresses high-dimensional LLM hidden states into the lower-dimensional memory space, while the decoder maps retrieved latent states back to the LLM's hidden size for processing.

\textbf{Cross-attention mechanism} \quad
To integrate new information, the memory slots act as Queries ($Q$), while the newly encoded experiences act as Keys ($K$) and Values ($V$). This allows the bank to selectively attend to relevant features of the current input.

\textbf{Gated update logic} \quad
To prevent memory drift and manage information overwrite, we implement a GRU-style update gate \citep{cho2014gru}.Let $M_{old}$ be the current state and $M_{attended}$ be the result of the cross-attention. The new state $M_{new}$ is calculated as:$$M_{new} = (1 - g) \odot M_{old} + g \odot M_{attended}$$where $g$ is the gate value produced by the $update\_gate$ network.

\subsection{System Dynamics: The Memory Loop}

The interaction between the LLM and the latent memory bank follows a three-stage execution cycle:

\textbf{Extraction} \quad The frozen LLM processes the input to generate raw hidden states.

\textbf{Retrieval} \quad The model queries the memory bank based on the encoded current hidden states. 

\textbf{Injection} \quad The retrieved latent information is decoded and concatenated with the original states through an gated injection layer, creating enhanced hidden states which are passed to the original LLM's language modeling head to produce logits.

\textbf{Consolidation} \quad The encoded original hidden states are fed back into the memory bank to update the memory slots via the cross-attention and gating mechanism, ensuring the memory evolves for the next interaction.

\subsection{Training Objective}

The memory system is trained using a composite loss function designed to optimize its performance on a primary task while encouraging desirable memory behaviors through regularization. The total loss $L_{total}$ is a weighted sum of three components:

\begin{enumerate}[label=\arabic*)]
    \item \textbf{Primary task loss} \quad The main objective is to train the memory module to improve the model's ability to predict the next token in a sequence. The enhanced hidden states are passed through the frozen LLM's language modeling head to produce memory-augmented logits $\hat{y}$. The primary loss $L_{CLM}$ is the standard cross-entropy loss between these predicted logits and the ground-truth target tokens $y$:
    $$L_{CLM} = -\frac{1}{T-1} \sum_{t=1}^{T-1} \log P(x_{t+1} \mid x_{1:t})$$
    
    \item \textbf{Sparsity loss} \quad An $L1$ penalty applied to encourage the model to use a sparse, focused set of memory slots, preventing it from storing redundant information:
    $$L_{sparsity} = \frac{1}{M} \sum \abs{s_i}$$
    where $s_i$ stands for each slot's importance score and $M$ is the total number of slots.
    
    \item \textbf{Entropy loss} \quad To prevent the model from relying on only one or two memory slots, we encourage diversity in memory usage by maximizing the entropy of the importance score distribution $p$, which is achieved by minimizing the negative entropy:
    $$L_{entropy} = \sum p_i \times \log(p_i)$$
    where $p$ is the softmax-normalized distribution of scores s.
\end{enumerate}

The final loss is a linear combination of the above components, weighted by hyperparameters $\lambda_s$ and $\lambda_e$:
$$  L_{total} = L_{CLM} + \lambda_s \times L_{sparsity} + \lambda_e \times L_{entropy}$$

\section{Experiments}

\subsection{Datasets}
We evaluate our memory-augmented architecture on two distinct challenges:

\begin{enumerate}[label=\arabic*)]
    \item \textbf{HotpotQA} \quad Evaluates multi-hop reasoning in a distractor setting.
    \item \textbf{ZsRE (Zero-Shot Relation Extraction)} \quad Evaluates the model's ability to extract and geralize factual relations.
\end{enumerate} 

For HotpotQA, we report Exact Match (EM) and F1 Score. For ZsRE, we report Accuracy, measuring the model's ability to correctly identify the object of a relation given a subject and a query.

\subsection{Baselines} 
We compare our memory-enhanced architecture against the standard vanilla versions of the respective base models:

\begin{enumerate}[label=\arabic*)]
    \item \textbf{GPT-2} \quad A small-scale baseline to test the module’s impact on limited parameter budgets.
    \item \textbf{Llama 3.1-8B} \quad A modern, high-capacity baseline to test scaling efficiency.
\end{enumerate}

\section{Results}

\subsection{Scaling Analysis}

The integration of the memory module consistently improves performance across both datasets and model scales.

\begin{table}[htbp]
  \centering
  \small    
  \begin{tabularx}{\textwidth}{l XXXXXX}
    \toprule
    & \multicolumn{2}{c}{\textbf{Answer}} & \multicolumn{2}{c}{\textbf{Sup Fact}} & \multicolumn{2}{c}{\textbf{Joint}} \\
    \cmidrule(lr){2-3} \cmidrule(lr){4-5} \cmidrule(lr){6-7}
    \textbf{Model} & \textbf{EM} & \textbf{F1} & \textbf{EM} & \textbf{F1} & \textbf{EM} & \textbf{F1} \\
    \midrule
    GPT-2 {\tiny(Vanilla)} & 35.47 & 45.52 & 15.18 & 51.84 & 11.35 & 30.72 \\
    \rowcolor{lightblue}
    GPT-2 {\tiny(G-MemLLM)} & 41.92 \up{6.45} & 54.08 \up{8.56} & 22.63 \up{7.45} & 60.17 \up{8.33} & 15.08 \up{3.73} & 38.51 \up{7.79} \\
    
    \addlinespace[1em]
    
    Llama 3.1-8B {\tiny(Vanilla)} & 68.53 & 79.27 & 62.19 & 76.53 & 51.43 & 72.15 \\
    \rowcolor{lightblue}
    Llama 3.1-8B {\tiny(G-MemLLM)} & 72.38 \up{3.85} & 82.12 \up{2.85} & 67.13 \up{4.94} & 83.42 \up{6.89} & 54.82 \up{3.39} & 78.23 \up{6.08} \\
    \bottomrule
  \end{tabularx}
  
  \caption{Performance on HotpotQA (Multi-hop Reasoning). Metrics include Answer, Supporting Facts (Sup Fact), and Joint evaluation of both.}
  \label{tab:hotpotqa}
\end{table}


\begin{table}[htbp]
  \centering
  \small    
  \begin{tabularx}{\textwidth}{l XXXXX}
    \toprule
    \textbf{Model} & \textbf{Score} & \textbf{Efficacy} & \textbf{Generalization} & \textbf{Specificity} \\
    \midrule
    Llama 3.1-8B {\tiny(Vanilla)}          & 55.63 & 55.92 & 54.71 & 56.31 \\
    \rowcolor{lightblue}
    Llama 3.1-8B{\tiny(G-MemLLM)}  & 63.03 \up{13.3\%} & 60.18 & 59.10 & 56.78 \\
    \bottomrule
  \end{tabularx}
  \captionsetup{justification=raggedright, singlelinecheck=false} 
  \caption{Evaluation results on the ZsRE dataset for different models.}
  \label{tab:zsre_results}
\end{table}

The empirical results summarized in Table 1 demonstrate that the G-MemLLM module consistently enhances performance across all metrics for both small-scale and large-scale models. For GPT-2, the addition of the latent memory bank yields a substantial improvement in reasoning capability, specifically boosting the Answer F1 score by 8.56 points (from 45.52 to 54.08) and Joint F1 by 7.79 points. Notably, the performance gains are not limited to smaller architectures; Llama 3.1-8B also shows significant progression. While the base Llama 3.1 model already exhibits strong performance, G-MemLLM further elevates its Supporting Fact (Sup Fact) F1 by 6.89 points (achieving 83.42) and its Joint F1 by 6.08 points. This trend suggests that G-MemLLM is particularly effective at evidence grounding, the ability to identify supporting facts, which is the primary bottleneck in multi-hop reasoning tasks like HotpotQA.

As shown in Table 2, the memory module significantly boosts performance on ZsRE. For Llama 3.1-8B, we observe a 13.3\% absolute increase in accuracy. This suggests that the memory module acts as a specialized buffer that helps the model disambiguate complex relations that are often conflated in the standard feed-forward layers.

A key finding of our experiments is the super-linear scaling benefit of the memory module:

\textbf{Small Scale (GPT-2)} \quad The memory module provides a baseline capability for multi-hop tasks that the vanilla model lacks, effectively acting as a scratchpad for intermediate steps.

\textbf{Large Scale (Llama 3.1-8B)} \quad As the model size increases to 8B, the memory module's utility shifts from basic retention to efficient indexing. The larger model uses the memory module to organize its vast internal knowledge more effectively, leading to the significant 13.3\% jump in ZsRE accuracy.

\begin{table}[htbp]
  \centering
  \small
  \begin{minipage}{0.48\textwidth}
    \centering
    \begin{tabularx}{\textwidth}{l XXX}
      \toprule
      \textbf{Slots} & \textbf{Score} & \textbf{$\Delta$(\%)} & \textbf{Over.} \\
      \midrule
      0 (Van.)        & 58.53 & —      & 1.00x \\
      512             & 61.72 & +5.45  & 1.05x \\
      \rowcolor{lightblue}
      1024 (Prop.)    & 63.03 & +2.12  & 1.12x \\
      2048            & 63.21 & +0.28  & 1.25x \\
      \bottomrule
    \end{tabularx}
    \captionof{table}{Ablation study on memory slot count ($S$) on ZsRE score.}
    \label{tab:memory_slots}
  \end{minipage}
  \hfill 
  \begin{minipage}{0.48\textwidth}
    \centering
    \includegraphics[width=\textwidth]{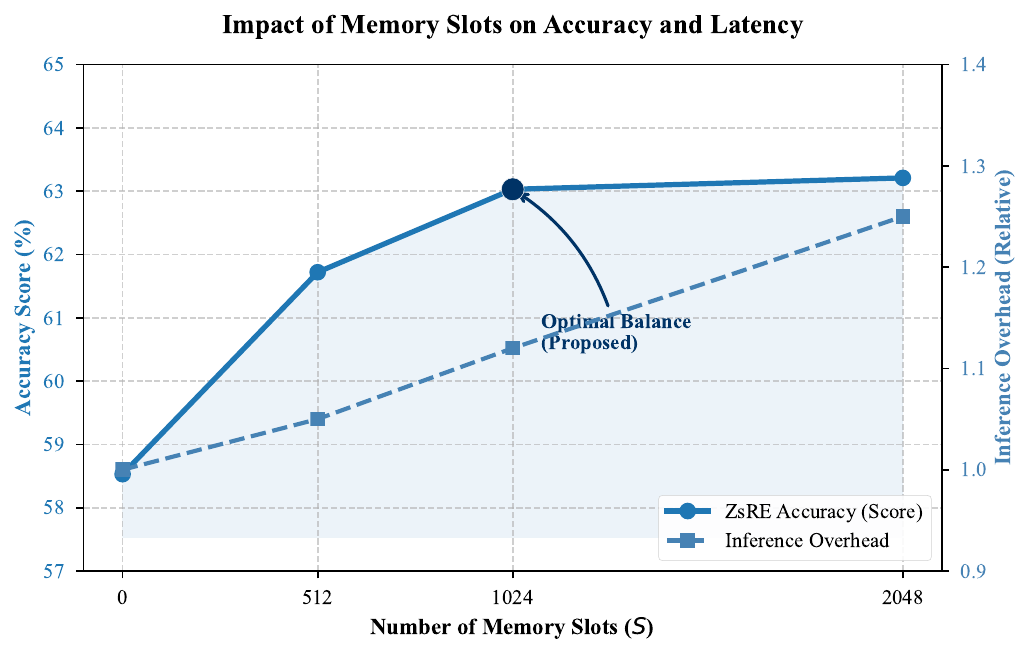}
    \captionof{figure}{Trend of performance gains and computational overhead.}
    \label{fig:memory_slots_plot}
  \end{minipage}
\end{table}

\subsection{Ablation Study: Memory Density}

We investigated how the size of the memory module affects performance on ZsRE using the Llama 3.1-8B backbone.

The ablation shows that 1024 slots provide the optimal balance between accuracy gains and computational overhead. Doubling the slots to 2048 provides diminishing returns (+0.28\%), suggesting a saturation point in relational storage for the 8B parameter scale.

\section{Summary}

This paper presents G-MemLLM, a memory-augmented architecture designed to enhance the multi-hop reasoning and relational knowledge retention of Large Language Models. By integrating a trainable latent memory bank with a frozen LLM backbone, we decouple linguistic processing from stateful information storage. Our primary technical innovation is the implementation of a GRU-style gated update logic, which allows the model to selectively preserve or overwrite latent memory slots. This mechanism, supported by a composite loss function emphasizing sparsity and entropy, successfully mitigates common issues such as context rot and information dilution found in traditional recurrent or compressed architectures.

Empirical evaluations across scales, from GPT-2 (124M) to Llama 3.1-8B, demonstrate that G-MemLLM provides significant performance gains on the HotpotQA and ZsRE benchmarks. These results suggest that as LLMs grow in capacity, they become increasingly efficient at using explicit memory modules to index and organize their internal knowledge.

As a machine learning course project, this work highlights the transition from passive context windows to active gated cognitive architectures. We successfully navigated challenges related to memory collapse and high-dimensional state projection on limited hardware. The findings underscore that adding a small, trainable working memory, representing less than 3\% additional parameters, is a highly efficient strategy for extending the reasoning capabilities of pre-trained models without the prohibitive cost of full-parameter fine-tuning.

\bibliography{iclr2025_conference}
\bibliographystyle{iclr2025_conference}

\end{document}

%% file: math_commands.tex
\usepackage{amsmath,amsfonts,bm}

\def\eqref#1{equation~\ref{#1}}

\def\1{\bm{1}}

\DeclareMathAlphabet{\mathsfit}{\encodingdefault}{\sfdefault}{m}{sl}
\SetMathAlphabet{\mathsfit}{bold}{\encodingdefault}{\sfdefault}{bx}{n}

